\documentclass[10pt,twocolumn,letterpaper]{article}

\usepackage{wacv}
\usepackage{times}
\usepackage{epsfig}
\usepackage{graphicx}
\usepackage{amsmath}
\usepackage{amssymb}
\usepackage{subcaption}
\usepackage{multirow}
\usepackage{float}
\usepackage{url}
\newcommand{\norm}[1]{\left\lVert#1\right\rVert}
\usepackage[shortcuts]{extdash}
\newcommand{\tableImHeight}{.69in}
\newcommand{\tableImHeightTwo}{.66in}
\newcommand{\tableImWidth}{.4}

\wacvfinalcopy 

\def\wacvPaperID{96} 

\ifwacvfinal\fi
\begin{document}

\title{Visualizing Deep Similarity Networks}

\author{Abby Stylianou \\
George Washington University\\
{\tt\small astylianou@gwu.edu}
\and
Richard Souvenir \\
Temple University\\
{\tt\small souvenir@temple.edu}
\and
Robert Pless \\
George Washington University\\
{\tt\small pless@gwu.edu}
}

\maketitle
\thispagestyle{empty}

\begin{abstract}
For convolutional neural network models that optimize an image embedding, we propose a method to highlight the regions of images that contribute most to pairwise similarity. 
This work is a corollary to the visualization tools developed for classification networks,
but applicable to the problem domains better suited to similarity learning. The visualization shows how similarity networks that are fine-tuned learn to focus on different features. We also generalize our approach to embedding networks that use different pooling strategies and provide a simple mechanism to support image similarity searches on objects or sub-regions in the query image.
\end{abstract}

\section{Introduction}
While convolutional neural networks have become a transformative tool for many image analysis tasks, it is still common in the literature to describe these deep learning approaches as ''black boxes''. To address these concerns, there have been substantial efforts to understand and visualize the features of classification networks~\cite{bau2017network,netdissect2017,visualization_techreport,deepInside,szegedy2015going,ZeilerF13,cam,scenecnn_iclr15}. However, much less work has focused on visualizing and understanding similarity networks, which learn an embedding that maps similar examples to nearby vectors in feature space and dissimilar examples to be far apart~\cite{Ahmed_2015_CVPR,RTC16,tolias2016rmac}. 


\begin{figure}[t]
    \setlength\tabcolsep{1.25pt}
    \centering
        \begin{tabular}{c|ccc}
             Query & \multicolumn{3}{|c}{Top Matches}  \\
             \includegraphics[height=\tableImHeight]{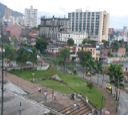}
             &
             \fcolorbox{green}{green}{\includegraphics[height=\tableImHeight]{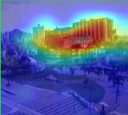}}
             &
             \fcolorbox{green}{green}{\includegraphics[height=\tableImHeight]{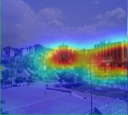}}
             &
             \fcolorbox{green}{green}{\includegraphics[height=\tableImHeight]{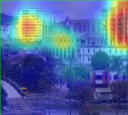}}
             \\
             \includegraphics[height=\tableImHeight]{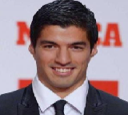}
             &
             \fcolorbox{green}{green}{\includegraphics[height=\tableImHeight]{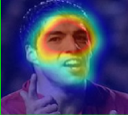}}
             &
             \fcolorbox{green}{green}{\includegraphics[height=\tableImHeight]{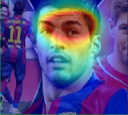}}
             &
             \fcolorbox{red}{red}{\includegraphics[height=\tableImHeight]{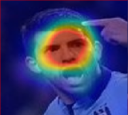}}
             \\
             \includegraphics[height=\tableImHeight]{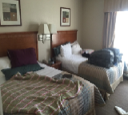}
             &
             \fcolorbox{green}{green}{\includegraphics[height=\tableImHeight]{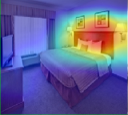}}
             &
             \fcolorbox{red}{red}{\includegraphics[height=\tableImHeight]{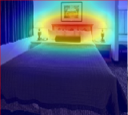}}
             &
             \fcolorbox{red}{red}{\includegraphics[height=\tableImHeight]{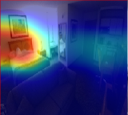}}
        \end{tabular}
     \caption{Our approach to visualizing the embeddings generated by deep similarity networks calculates the contribution of each pixel location to the overall similarity between two images. We evaluate our approach on a variety of problem domains and network architectures.}
     \label{fig:frontPage}
\end{figure}

Our approach highlights the image regions that contributed the most to the overall similarity between two images. Figure~\ref{fig:frontPage} shows example visualizations for the top image retrieval results from three different application domains (Google Landmarks~\cite{googleLandmarks}, VGG-Faces~\cite{vggfaces}, and Traffickcam Hotel Rooms~\cite{aipr2015}). Each row of the figure shows a query image and the three most similar database images returned from a network trained for the respective task. The heatmap overlay shows the relative spatial contribution of each image to the similarity score with the query.
 
Our approach aligns with the recent trend toward explainability for learning-based tasks and extends recent work in visualizing classification networks to the case of similarity networks. Our specific contributions include:
\begin{itemize}
\itemsep0em 
    \item a novel visualization approach for similarity networks;
    \item an analysis of the effect of training and ``late-stage'' pooling strategies for similarity networks; and,
    \item an approach to using similarity visualizations to support object- and region-based image retrieval.
\end{itemize}



\section{Background}
Visualizations provide a way to better understand the learning process underlying deep neural networks. Much of the work in this area focuses on visualizations for classification networks and not similarity networks. While networks used for each type of problem share many similarities, the differences in the output (i.e., sparse vs. dense feature vectors) is significant, requiring new methods for visualizing similarity networks.

\paragraph{CNN Visualization}
Previous work on CNN visualizations can be broadly categorized by
the depth of the portion of the networked being visualized. Some methods provide visualizations that highlight the inner layer activations~\cite{netdissect2017,visualization_techreport,scenecnn_iclr15}. A majority of the work targets the output layer to produce visualizations which seek to explain
why classification networks output a particular label for an image. These include approaches that mask off parts of the input images and provide a visual quantification of the impact on the output classification~\cite{ZeilerF13}. Another approach generates saliency maps, which represent which pixels in an image contributed to a particular output node~\cite{deepInside}. There has been work that generates class activation maps, which map an output back to the last convolutional layer in the network by weighting the filters in that layer by the weights between the final pooling layer and the output feature~\cite{cam}. Inception~\cite{szegedy2015going}, which hallucinates images that activate a particular class from random noise, can also serve as visualization tool to provide insight into the learning process. 

\paragraph{Similarity Learning}
Much of the work in similarity learning with deep neural networks focuses on learning better similarity functions using, for example, pairwise losses~\cite{sun2014deep,wang2014learning,yi2014deep}, triplet losses~\cite{HermansBeyer2017Arxiv,schroff2015facenet,song2016deep,ustinova2016learning}, and direct embedding~\cite{proxy}. Compared to the efforts toward understanding classification networks, there has been much less work in visualizing and analyzing similarity networks. One method visualizes the similarity of single filters from the different convolutional layers of an embedding network~\cite{Ahmed_2015_CVPR}. Another method computes image similarity as the inner product between the normalized elements of a final max pooling layer and produces a visualization with bounding boxes around highly active regions for the ten features that contribute most to the similarity of a pair of images~\cite{RTC16,tolias2016rmac}. 

Visualizing a few features is effective for networks that tend to be sparse, but in Section~\ref{sec:FeatureImportance} we show that in embedding networks the similarity tends to be explained by a large number of features.  This motivates our approach to visualize how all features affect the similarity score.

\begin{figure*}
    \centering
    \includegraphics[width=0.99\textwidth]{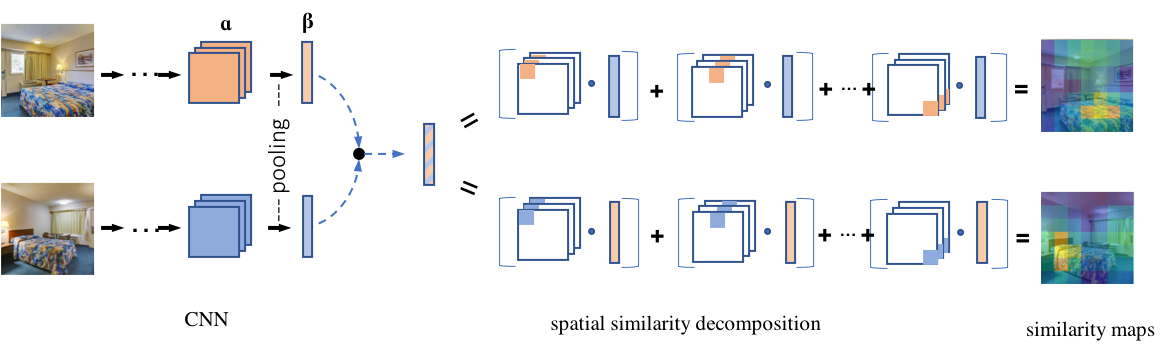}
    \caption{Our approach considers similarity networks with a final convolutional layer, $\boldsymbol{\alpha}$, followed by a pooling operation which produces output features, $\boldsymbol{\beta}$. Similarity between two images is measured as the dot product of these output features after normalization.  Factoring this value produces visualizations that highlight how much each region of the image contributes to the similarity.}
    \label{fig:visApproach}
\end{figure*}
\section{Visualization Approach}
\label{sec:model}

Networks used in similarity learning broadly consist of: (1) a convolutional portion, (2) a "flattening" operation (usually max or global average pooling), and (3) a fully-connected portion. A recent study covering a number of image retrieval tasks, however, suggests that the best generalization performance is obtained using the output from the layer immediately after the pooling operation~\cite{vo2018generalization}. Our approach is applicable to networks of this structure, including popular models such as the Resnet~\cite{resnet} and VGG~\cite{vggfaces} network architectures.  

Given an input image, $I$, and a trained similarity network, our approach relies on the activations of the layers before and after the pooling operation. Let $\boldsymbol{\alpha}$ represent the $K \times K \times C$ tensor of the last convolutional layer, where $K$ represents the length and width (usually equal) and $C$ represents the number of filters. Let $\boldsymbol{\beta}$ represent the $C$-dimensional vector after the pooling operation for an image, as shown in Figure~\ref{fig:visApproach}. In similarity learning, the dot product of these normalized feature vectors is a widely-used similarity function~\cite{bell2015learning,proxy,schroff2015facenet,sun2014deep,wang2014learning,yi2014deep}, so the similarity of two images $I^{(i)},I^{(j)}$ can be written as:
\begin{equation}
s(\boldsymbol{\beta}^{(i)},\boldsymbol{\beta}^{(j)}) = \frac{\boldsymbol{\beta}^{(i)} \cdot \boldsymbol{\beta}^{(j)}}{\norm{\boldsymbol{\beta}^{(i)}} \norm{\boldsymbol{\beta}^{(j)}}}
\label{eq:cosineSim}
\end{equation}
Our visualization approach results in spatial similarity maps, where the overall similarity between two image feature vectors is spatially decomposed to highlight the contribution of image regions to the overall pairwise similarity, as shown in Figure~\ref{fig:visApproach}. Computing the similarity maps depends on the flattening operation between the convolutional portion of the network and the output feature. Max pooling and global average pooling are the most commonly applied operations at this stage in modern networks. We show how our similarity maps are computed for each case.

\subsection{Average Pooling}
For networks which employ average pooling as the flattening operation, the output feature, $\boldsymbol{\beta}$, is:
\begin{equation} 
\boldsymbol{\beta} = \frac{1}{K^2}\sum_{x,y} \boldsymbol{\alpha}_{(x,y)}
\label{eq:avgPool}
\end{equation}
\noindent where $\boldsymbol{\alpha}_{(x,y)}$ represents the $C$-dimensional slice of $\boldsymbol{\alpha}$  at spatial location $(x,y)$. The similarity of images $I^{(i)}$ and $I^{(j)}$ can be directly decomposed spatially, by substituting $\boldsymbol{\beta}^{(i)}$ in Equation~\ref{eq:cosineSim} with Equation~\ref{eq:avgPool}:
\begin{align}
    s(\boldsymbol{\beta}^{(i)},\boldsymbol{\beta}^{(j)}) &= \frac{\boldsymbol{\beta}^{(i)}\cdot\boldsymbol{\beta}^{(j)}}{\norm{\boldsymbol{\beta}^{(i)}} \norm{\boldsymbol{\beta}^{(j)}}}\nonumber\\
    &=
    \frac{\frac{1}{K^2}\left(\boldsymbol{\alpha}^{(i)}_{(1,1)} + \ldots + \boldsymbol{\alpha}^{(i)}_{(K,K)}\right)  \cdot \boldsymbol{\beta^{(j)}}}{\norm{\boldsymbol{\beta}^{(i)}} \norm{\boldsymbol{\beta}^{(j)}}}\nonumber\\
    &= \frac{\boldsymbol{\alpha}^{(i)}_{(1,1)} \cdot \boldsymbol{\beta}^{(j)} + \ldots + \boldsymbol{\alpha}^{(i)}_{(K,K)} \cdot \boldsymbol{\beta}^{(j)}
    }{Z}
\end{align}
\noindent where $Z$ is the normalizing factor $K^2 \norm{\boldsymbol{\beta}^{(i)}} \norm{\boldsymbol{\beta}^{(j)}}$.

These terms can be rearranged spatially and visualized as a heat-map to show the relative contribution of each part of the image to the overall similarity. Symmetrically, the similarity can be decomposed to highlight the contribution of the other image in the pair to the overall similarity, as shown on the right side of Figure~\ref{fig:visApproach}.

\subsection{Max Pooling}
With a modification, the approach can also accommodate networks that use max pooling as the flattening operation. In max pooling, each element of an output vector $\boldsymbol{\beta}$ is equal to the max value of the activation of its corresponding filter in the last convolutional layer:
\begin{equation}
\boldsymbol{\beta} = \max_{x,y} \boldsymbol{\alpha}_{(x,y)}
\label{eq:maxPool}
\end{equation}
\noindent Unlike average pooling, where each of the composite components contribute equally to the output feature, decomposing max pooled features requires an additional step. For a max pooled feature, $\boldsymbol{\beta}$, we construct a surrogate tensor, $\boldsymbol{\hat{\alpha}}$, for the convolutional portion as follows:
\begin{equation}
    \hat{\alpha}_{(x,y,c)} = \begin{cases}
    0 & \text{if } \alpha_{(x,y,c)} \neq \beta_{(c)} \\
    \frac{\alpha_{(x,y,c)}}{N_{(c)}} & \text{if } \alpha_{(x,y,c)} = \beta_{(c)} \\
    \end{cases}
\end{equation}
\noindent where $N_{(c)}$ represents the number of spatial locations equal to the maximum value for filter $c$. That is, for each filter, we assign the maximum value to the location that generated it (divided evenly in cases of ties), and zero otherwise. This gives the following formulation for the spatial similarity decomposition in the case of max pooling:
\begin{align}
     s(\boldsymbol{\beta}^{(i)},\boldsymbol{\beta}^{(j)}) &=
    \frac{\boldsymbol{\hat{\alpha}}^{(i)}_{(1,1)} \cdot \boldsymbol{\hat{\beta}}^{(j)} + \ldots + \boldsymbol{\hat{\alpha}}^{(i)}_{(K,K)} \cdot \boldsymbol{\hat{\beta}}^{(j)}}{\norm{\boldsymbol{\beta}^{(i)}}\norm{\boldsymbol{\beta}^{(j)}}}
\end{align}
Similar to the case for average pooling, similarity maps can be computed in either direction for a pair of images. 

We scale the heatmaps using bilinear interpolation and blend them with the original image to show which parts of the images contribute to the similarity scores.

\section{Results}
Similarity networks trained for three different problem domains are used to test the approach. Except where noted, we use the following network architectures and output features. For the Google Landmarks~\cite{googleLandmarks} and TraffickCam Hotel Rooms~\cite{aipr2015} datasets, we fine-tune a Resnet-50~\cite{resnet} network from pre-trained ILSVRC weights~\cite{ILSVRC15} using the combinatorial variant of triplet loss described in~\cite{HermansBeyer2017Arxiv}. For the VGG-Faces dataset, we use the VGG-Faces network trained on the VGG-Faces2 dataset~\cite{vggface2,vggfaces}. For each of the networks, we use the layer immediately after the pooling operation as our output features (2048-D for Resnet-50, and 512-D for VGG-Faces).

\begin{figure}
    \centering
    \includegraphics[width=0.9\columnwidth]{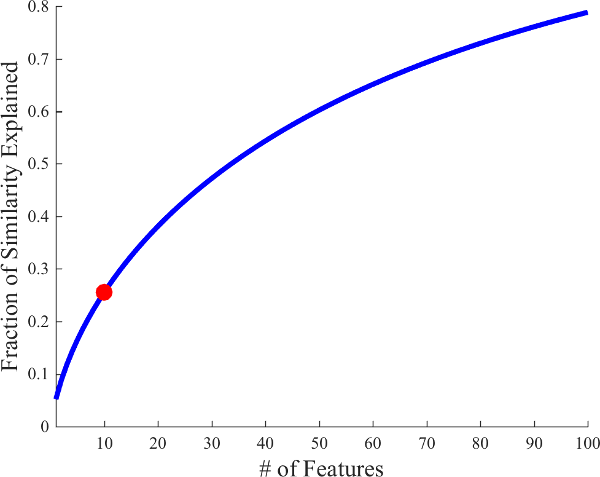}
    \caption{The plot shows the average contribution of the top\==K components of the feature vectors to the similarity score between pairs of images from the same class using the pre-trained VGG-Faces dataset. The top 10 features (the number of features visualized in prior work, and identified in this plot by a red dot), account for less than 30\% of the similarity score between two images, motivating our attempt to visualize all features.}
    \label{fig:contribByFeature}
\end{figure}

\subsection{Feature Importance}
\label{sec:FeatureImportance}
Prior work in understanding similarity networks focuses on either a few filters or few regions that contribute most to the similarity between a pair of images~\cite{Ahmed_2015_CVPR,RTC16,tolias2016rmac}. Our visualization approach, by comparison, summarizes the contribution of every feature to the similarity between a pair of images. 

In the following experiment, we demonstrate that for similarity networks, the top few most important components only represent a small fraction of the overall image similarity.  Figure~\ref{fig:contribByFeature} shows the average contribution of the first $k$ components for 1000 randomly sampled pairs of images from the same class using the pre-trained VGG-Faces network. The top 10 features (the number of features visualized in prior work, and identified in this plot by a red dot) contribute less than 30\% of the overall similarity score. This suggests that, unlike classification networks which output sparse feature vectors, understanding the output of similarity networks requires a visualization approach that explains more than only a few features at once. Our approach to visualizing similarity networks incorporates all of the feature vector components and calculates the contribution of each pixel location to the overall similarity between two images.

\begin{figure}[t]
    \centering
    \begin{subfigure}[b]{\columnwidth}
        \centering
        \begin{tabular}{cc}
            \includegraphics[width=\tableImWidth\columnwidth]{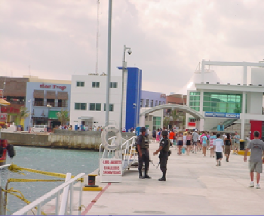} &  
            \includegraphics[width=\tableImWidth\columnwidth]{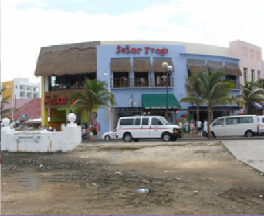}\\
            \includegraphics[width=\tableImWidth\columnwidth]{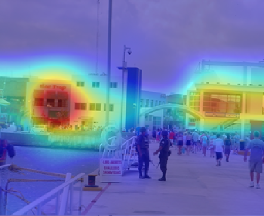} &  
            \includegraphics[width=\tableImWidth\columnwidth]{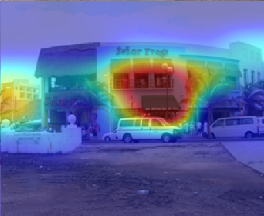}\\
        \end{tabular}
    \end{subfigure}
    \\\vspace{7px}
    \begin{subfigure}[b]{\columnwidth}
        \centering
        \begin{tabular}{cc}
            \includegraphics[width=\tableImWidth\columnwidth]{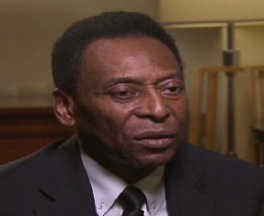} &  
            \includegraphics[width=\tableImWidth\columnwidth]{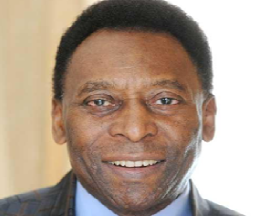}\\
            \includegraphics[width=\tableImWidth\columnwidth]{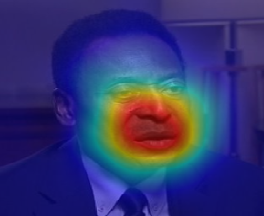} &  
            \includegraphics[width=\tableImWidth\columnwidth]{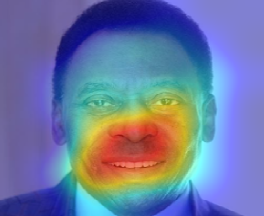}
        \end{tabular}
    \end{subfigure}
    \\\vspace{7px}
    \begin{subfigure}[b]{\columnwidth}
        \centering
        \begin{tabular}{cc}
            \includegraphics[width=\tableImWidth\columnwidth]{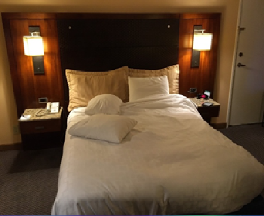} &  
            \includegraphics[width=\tableImWidth\columnwidth]{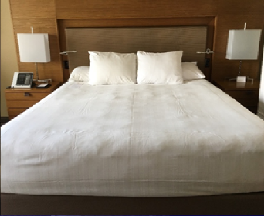}\\
            \includegraphics[width=\tableImWidth\columnwidth]{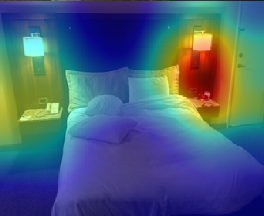} &  
            \includegraphics[width=\tableImWidth\columnwidth]{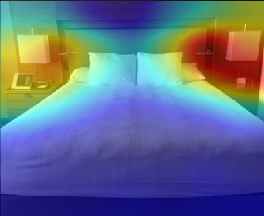}
        \end{tabular}
    \end{subfigure}
    \caption{Visualizations to understand image similarity. (Top) For two images of the same landmark, the visualization highlights the building in the background in left image, but the foreground in the right. (Middle) For two images of the same person, the nose and mouth region are highlighted. (Bottom) For two images of rooms from different hotels, the visualization highlights the similar light fixtures mounted to the headboard.}
    \label{fig:clarification}
\end{figure}

\subsection{Visualizing Pairwise Similarity}
Figure~\ref{fig:clarification} shows pairs of images that produced high similarity scores. In the top pair of images from the Google Landmarks dataset, the viewpoints are quite different, but the visualization approach highlights the specific building that the network identified as being similar. This building is in the foreground of one of the images, but hidden in the background of the other. The middle pair of images are of the same gentleman in the VGG-Faces dataset. the visualization highlights his lower facial features. The final pair of images is from different hotels in the TraffickCam Hotel Rooms dataset. The visualization highlights that both rooms have similar light fixtures mounted to the headboard. These examples demonstrate the ability of the visualization approach in explaining why a network produces similar embeddings for a pair of images, even in cases where that may not be readily apparent to a human observer looking at the images.

\begin{figure*}
    \centering
    \setlength\tabcolsep{1.25pt}
    \begin{subfigure}[b]{.47\textwidth}
        \begin{tabular}{c|ccc}
             Query & \multicolumn{3}{|c}{Top Matches}  \\
             \includegraphics[height=\tableImHeight]{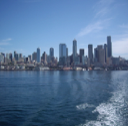}
             &
             \fcolorbox{green}{green}{\includegraphics[height=\tableImHeight]{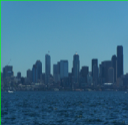}}
             &
             \fcolorbox{green}{green}{\includegraphics[height=\tableImHeight]{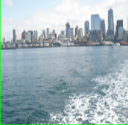}}
             &
             \fcolorbox{green}{green}{\includegraphics[height=\tableImHeight]{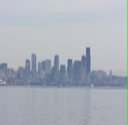}}
             \\
             Initial
             &
             \raisebox{-.5\height}{\includegraphics[height=\tableImHeight]{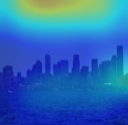}}
             &
             \raisebox{-.5\height}{\includegraphics[height=\tableImHeight]{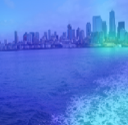}}
             &
             \raisebox{-.5\height}{\includegraphics[height=\tableImHeight]{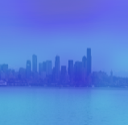}}
             \\
             5k iterations
             &
             \raisebox{-.5\height}{\includegraphics[height=\tableImHeight]{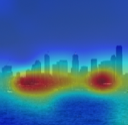}}
             &
             \raisebox{-.5\height}{\includegraphics[height=\tableImHeight]{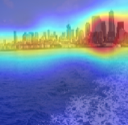}}
             &
             \raisebox{-.5\height}{\includegraphics[height=\tableImHeight]{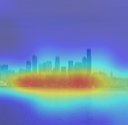}}
             \\
             25k iterations
             &
             \raisebox{-.5\height}{\includegraphics[height=\tableImHeight]{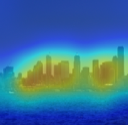}}
             &
             \raisebox{-.5\height}{\includegraphics[height=\tableImHeight]{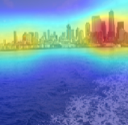}}
             &
             \raisebox{-.5\height}{\includegraphics[height=\tableImHeight]{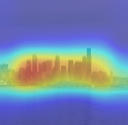}}
             \\
             50k iterations
             &
             \raisebox{-.5\height}{\includegraphics[height=\tableImHeight]{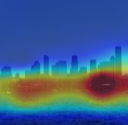}}
             &
             \raisebox{-.5\height}{\includegraphics[height=\tableImHeight]{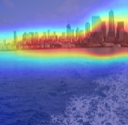}}
             &
             \raisebox{-.5\height}{\includegraphics[height=\tableImHeight]{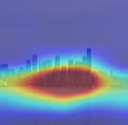}}
        \end{tabular}
    \end{subfigure}
    \hspace{1em}
    \begin{subfigure}[b]{.47\textwidth}
    \begin{tabular}{c|ccc}
             Query & \multicolumn{3}{|c}{Top Matches}  \\
             \includegraphics[height=\tableImHeight]{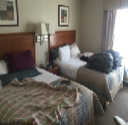}
             &
             \fcolorbox{green}{green}{\includegraphics[height=\tableImHeight]{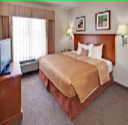}}
             &
             \fcolorbox{red}{red}{\includegraphics[height=\tableImHeight]{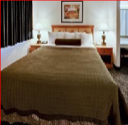}}
             &
             \fcolorbox{red}{red}{\includegraphics[height=\tableImHeight]{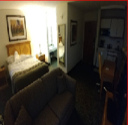}}
             \\
             Initial
             &
             \raisebox{-.5\height}{\includegraphics[height=\tableImHeight]{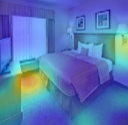}}
             &
             \raisebox{-.5\height}{\includegraphics[height=\tableImHeight]{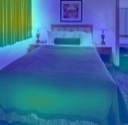}}
             &
             \raisebox{-.5\height}{\includegraphics[height=\tableImHeight]{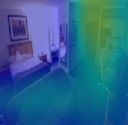}}
             \\
             5k iterations
             &
             \raisebox{-.5\height}{\includegraphics[height=\tableImHeight]{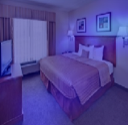}}
             &
             \raisebox{-.5\height}{\includegraphics[height=\tableImHeight]{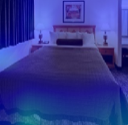}}
             &
             \raisebox{-.5\height}{\includegraphics[height=\tableImHeight]{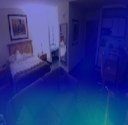}}
             \\
             50k iterations
             &
             \raisebox{-.5\height}{\includegraphics[height=\tableImHeight]{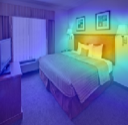}}
             &
             \raisebox{-.5\height}{\includegraphics[height=\tableImHeight]{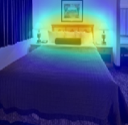}}
             &
             \raisebox{-.5\height}{\includegraphics[height=\tableImHeight]{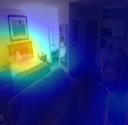}}
             \\
             100k iterations
             &
             \raisebox{-.5\height}{\includegraphics[height=\tableImHeight]{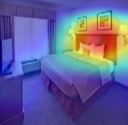}}
             &
             \raisebox{-.5\height}{\includegraphics[height=\tableImHeight]{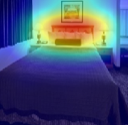}}
             &
             \raisebox{-.5\height}{\includegraphics[height=\tableImHeight]{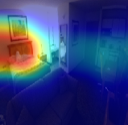}}
        \end{tabular}
    \end{subfigure}
    \caption{Each subfigure shows visualizations from networks pre-trained on ImageNet and fine-tuned on Google Landmarks (left) and TraffickCam Hotel Rooms (right) during the training process.}
    \label{fig:overTime}
\end{figure*}
\setlength\tabcolsep{2pt}
\begin{figure*}
    \centering
    \begin{subfigure}[b]{\columnwidth}
    \begin{tabular}{c|ccc}
             Query & \multicolumn{3}{|c}{Top Matches}  \\
             \includegraphics[height=\tableImHeight]{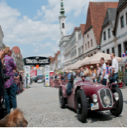}
             &
             \fcolorbox{green}{green}{\includegraphics[height=\tableImHeight]{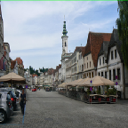}}
             &
             \fcolorbox{green}{green}{\includegraphics[height=\tableImHeight]{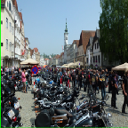}}
             &
             \fcolorbox{green}{green}{\includegraphics[height=\tableImHeight]{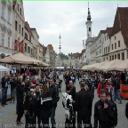}}
             \\
             From Scratch
             &
             \raisebox{-.5\height}{\includegraphics[height=\tableImHeight]{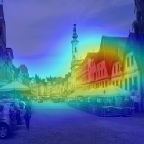}}
             &
             \raisebox{-.5\height}{\includegraphics[height=\tableImHeight]{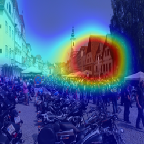}}
             &
             \raisebox{-.5\height}{\includegraphics[height=\tableImHeight]{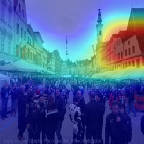}}
             \\
             Fine-tuned
             &
             \raisebox{-.5\height}{\includegraphics[height=\tableImHeight]{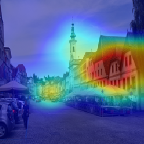}}
             &
             \raisebox{-.5\height}{\includegraphics[height=\tableImHeight]{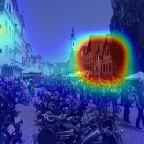}}
             &
             \raisebox{-.5\height}{\includegraphics[height=\tableImHeight]{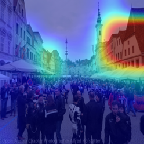}}
        \end{tabular}
    \end{subfigure}
    \hspace{.63cm}
    \begin{subfigure}[b]{\columnwidth}
    \begin{tabular}{c|ccc}
            Query & \multicolumn{3}{|c}{Top Matches}  \\
             \includegraphics[height=\tableImHeight]{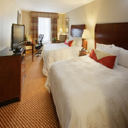}
             &
             \fcolorbox{green}{green}{\includegraphics[height=\tableImHeight]{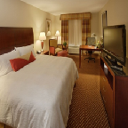}}
             &
             \fcolorbox{green}{green}{\includegraphics[height=\tableImHeight]{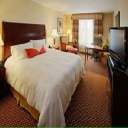}}
             &
             \fcolorbox{green}{green}{\includegraphics[height=\tableImHeight]{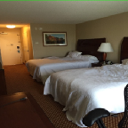}}
             \\
             From Scratch
             &
             \raisebox{-.5\height}{\includegraphics[height=\tableImHeight]{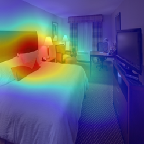}}
             &
             \raisebox{-.5\height}{\includegraphics[height=\tableImHeight]{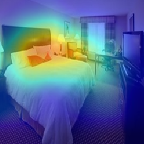}}
             &
             \raisebox{-.5\height}{\includegraphics[height=\tableImHeight]{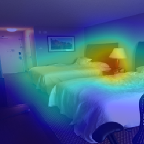}}
             \\
             Fine-tuned
             &
             \raisebox{-.5\height}{\includegraphics[height=\tableImHeight]{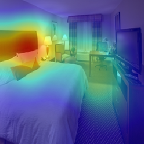}}
             &
             \raisebox{-.5\height}{\includegraphics[height=\tableImHeight]{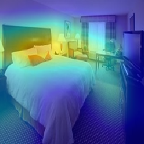}}
             &
             \raisebox{-.5\height}{\includegraphics[height=\tableImHeight]{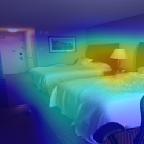}}
        \end{tabular}
    \end{subfigure}
    \caption{Fine-tuning vs. Training from Scratch. The visualization highlights that, regardless of the initialization, the networks converge to similar representations.}
    \label{fig:finetuning_vs_fromScratch}
\end{figure*}

\setlength\tabcolsep{2pt}
\begin{figure*}
    \centering
    \begin{subfigure}[b]{\columnwidth}
        \begin{tabular}{c|ccc}
             Query & \multicolumn{3}{|c}{Top Matches}  \\
             \includegraphics[height=\tableImHeightTwo]{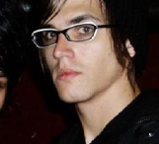}
             &
             \fcolorbox{green}{green}{\includegraphics[height=\tableImHeightTwo]{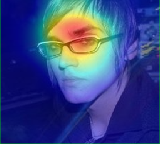}}
             &
             \fcolorbox{red}{red}{\includegraphics[height=\tableImHeightTwo]{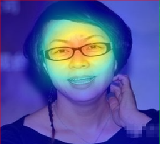}}
             &
             \fcolorbox{red}{red}{\includegraphics[height=\tableImHeightTwo]{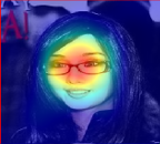}}
             \\
             \includegraphics[height=\tableImHeightTwo]{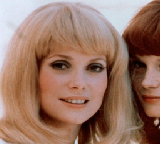}
             &
             \fcolorbox{red}{red}{\includegraphics[height=\tableImHeightTwo]{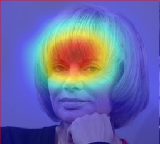}}
             &
             \fcolorbox{red}{red}{\includegraphics[height=\tableImHeightTwo]{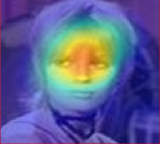}}
             &
             \fcolorbox{red}{red}{\includegraphics[height=\tableImHeightTwo]{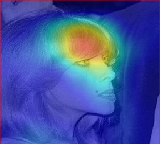}}
             \\
             \includegraphics[height=\tableImHeightTwo]{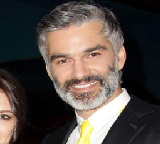}
             &
             \fcolorbox{green}{green}{\includegraphics[height=\tableImHeightTwo]{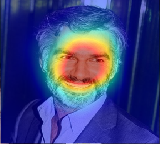}}
             &
             \fcolorbox{green}{green}{\includegraphics[height=\tableImHeightTwo]{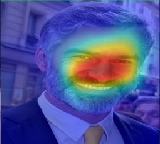}}
             &
             \fcolorbox{red}{red}{\includegraphics[height=\tableImHeightTwo]{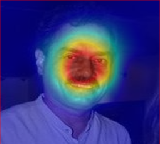}}
        \end{tabular}
        \caption{Average Pooling}
    \end{subfigure}
    \begin{subfigure}[b]{\columnwidth}
           \begin{tabular}{c|ccc}
           Query & \multicolumn{3}{|c}{Top Matches}  \\
             \includegraphics[height=\tableImHeightTwo]{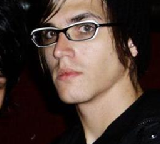}
             &
             \fcolorbox{green}{green}{\includegraphics[height=\tableImHeightTwo]{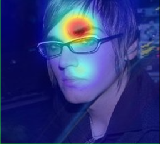}}
             &
             \fcolorbox{red}{red}{\includegraphics[height=\tableImHeightTwo]{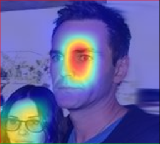}}
             &
             \fcolorbox{green}{green}{\includegraphics[height=\tableImHeightTwo]{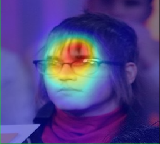}}
             \\
             \includegraphics[height=\tableImHeightTwo]{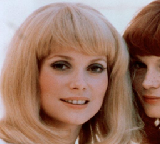}
             &
             \fcolorbox{red}{red}{\includegraphics[height=\tableImHeightTwo]{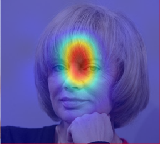}}
             &
             \fcolorbox{red}{red}{\includegraphics[height=\tableImHeightTwo]{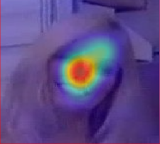}}
             &
             \fcolorbox{red}{red}{\includegraphics[height=\tableImHeightTwo]{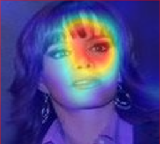}}
             \\
             \includegraphics[height=\tableImHeightTwo]{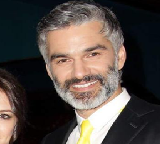}
             &
             \fcolorbox{green}{green}{\includegraphics[height=\tableImHeightTwo]{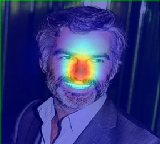}}
             &
             \fcolorbox{green}{green}{\includegraphics[height=\tableImHeightTwo]{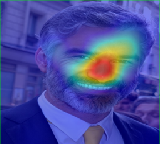}}
             &
             \fcolorbox{red}{red}{\includegraphics[height=\tableImHeightTwo]{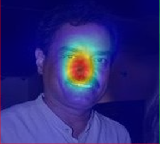}}
        \end{tabular}
        \caption{Max Pooling}
    \end{subfigure}
    \caption{Average vs. Max Pooling. For the same VGG-Faces network architecture, these visualizations show the pairwise similarity for models trained with average pooling and max pooling.}
    \label{fig:avg_vs_max_pooling}
\end{figure*}

\subsection{Similarity Learning During Training}
Figure~\ref{fig:overTime} shows the visualization for a query image and its top 3 most similar images during the training process. For the Google Landmarks dataset, we see that even by 5,000 iterations, the network has largely learned that it is 
the skyline that makes this scene recognizable. In subsequent iterations, the network refines the similarity metric and focuses on more specific regions, such as the buildings in the scene. On the TraffickCam Hotel Rooms dataset, on the other hand, the network takes longer to learn a similarity embedding. At 5,000 iterations, the network has not yet focused on specific elements of the hotel rooms. By 50,000 iterations, it becomes clear that the headboard is the relevant part of this particular set of images, and by 100,000 iterations, the network appears to be refining that focus. These examples demonstrate the utility of the visualization in understanding \textbf{\textit{when}} a network has learned a useful similarity metric, in addition to understanding what components of a scene the network has learned to focus on.

Another consideration when training similarity networks is whether to train from scratch or fine-tune from pre-trained weights. Figure~\ref{fig:finetuning_vs_fromScratch} shows the top three results for a query image, the similarity visualizations when trained from scratch, and when fine-tuned from pre-trained weights. In the examples from the Google Landmarks and TraffickCam Hotel Rooms dataset, we see that both the fine-tuned network and the network trained from scratch converged to similar encodings of similarity (e.g., both the fine-tuned network and network trained from scratch highlight the building facade in the Google Landmarks scene and the headboard in the TraffickCam hotel). These results suggest
that both approaches converge to features that encode the same important elements of the scenes and that it is reasonable to fine-tune from pre-trained weights (even from a fairly dissimilar task, such as a classification task trained on ILSVRC).

\subsection{Average vs. Max Pooling}

As described in Section~\ref{sec:model}, the visualization approach is applicable to networks with either average and max pooling at the end of the convolutional portion of the network. Figure~\ref{fig:avg_vs_max_pooling} shows the comparison between two VGG-Faces networks, one trained with average pooling and one with max pooling. 
For the same image pairs, the embeddings highlight different regions. For example, 
the average pooling network focuses on glasses in the first query image, while the max pooling network focuses more on eyebrow shape. Additionally, the regions of similarity are larger in the average pooling network compared to the max pooling network. This is reasonable as all of the regions contribute to output embedding in average pooling, but not max pooling.

\setlength\tabcolsep{2.5pt}
\renewcommand{\arraystretch}{3.25}
\begin{figure}
    \centering
    \begin{subfigure}[b]{\columnwidth}
        \begin{tabular}{ccccc}
            Class 1 &
            \raisebox{-.5\height}{\includegraphics[height=\tableImHeight]{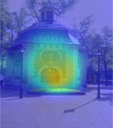}} &
            \raisebox{-.5\height}{\includegraphics[height=\tableImHeight]{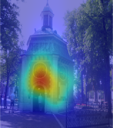}} &
            \raisebox{-.5\height}{\includegraphics[height=\tableImHeight]{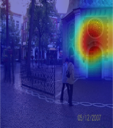}} &
            \raisebox{-.5\height}{\includegraphics[height=\tableImHeight]{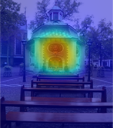}}
            \\
            Class 2 &
            \raisebox{-.5\height}{\includegraphics[height=\tableImHeight]{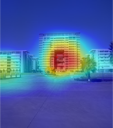}} &
            \raisebox{-.5\height}{\includegraphics[height=\tableImHeight]{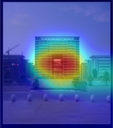}} &
            \raisebox{-.5\height}{\includegraphics[height=\tableImHeight]{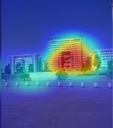}} &
            \raisebox{-.5\height}{\includegraphics[height=\tableImHeight]{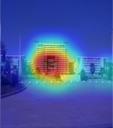}}
            \\
            Class 3 &
            \raisebox{-.5\height}{\includegraphics[height=\tableImHeight]{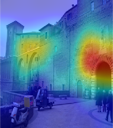}} &
            \raisebox{-.5\height}{\includegraphics[height=\tableImHeight]{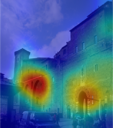}} &
            \raisebox{-.5\height}{\includegraphics[height=\tableImHeight]{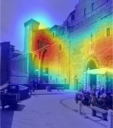}} &
            \raisebox{-.5\height}{\includegraphics[height=\tableImHeight]{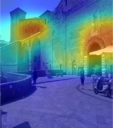}}
        \end{tabular}
        \caption{Google Landmarks}
    \end{subfigure}
    \begin{subfigure}[b]{\columnwidth}
        \begin{tabular}{ccccc}
            Class 1 &
            \raisebox{-.5\height}{\includegraphics[height=\tableImHeight]{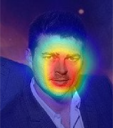}} &
            \raisebox{-.5\height}{\includegraphics[height=\tableImHeight]{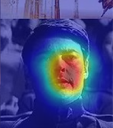}} &
            \raisebox{-.5\height}{\includegraphics[height=\tableImHeight]{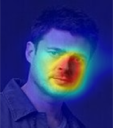}} &
            \raisebox{-.5\height}{\includegraphics[height=\tableImHeight]{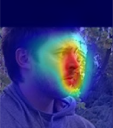}}
            \\
            Class 2 &
            \raisebox{-.5\height}{\includegraphics[height=\tableImHeight]{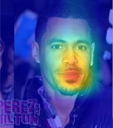}} &
            \raisebox{-.5\height}{\includegraphics[height=\tableImHeight]{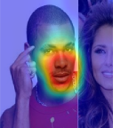}} &
            \raisebox{-.5\height}{\includegraphics[height=\tableImHeight]{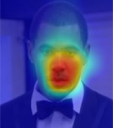}} &
            \raisebox{-.5\height}{\includegraphics[height=\tableImHeight]{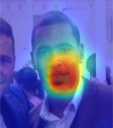}}
            \\
            Class 3 &
            \raisebox{-.5\height}{\includegraphics[height=\tableImHeight]{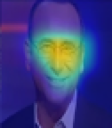}} &
            \raisebox{-.5\height}{\includegraphics[height=\tableImHeight]{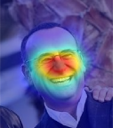}} &
            \raisebox{-.5\height}{\includegraphics[height=\tableImHeight]{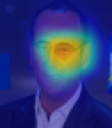}} &
            \raisebox{-.5\height}{\includegraphics[height=\tableImHeight]{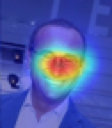}}
        \end{tabular}
        \caption{VGG-Faces}
    \end{subfigure}
    \begin{subfigure}[b]{\columnwidth}
        \begin{tabular}{ccccc}
            Class 1 &
            \raisebox{-.5\height}{\includegraphics[height=\tableImHeight]{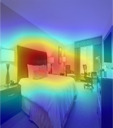}} &
            \raisebox{-.5\height}{\includegraphics[height=\tableImHeight]{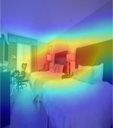}} &
            \raisebox{-.5\height}{\includegraphics[height=\tableImHeight]{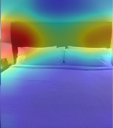}} &
            \raisebox{-.5\height}{\includegraphics[height=\tableImHeight]{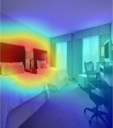}}
            \\
            Class 2 &
            \raisebox{-.5\height}{\includegraphics[height=\tableImHeight]{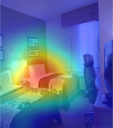}} &
            \raisebox{-.5\height}{\includegraphics[height=\tableImHeight]{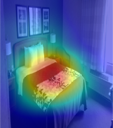}} &
            \raisebox{-.5\height}{\includegraphics[height=\tableImHeight]{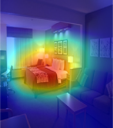}} &
            \raisebox{-.5\height}{\includegraphics[height=\tableImHeight]{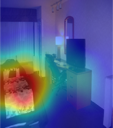}}
            \\
            Class 3 &
            \raisebox{-.5\height}{\includegraphics[height=\tableImHeight]{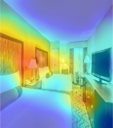}} &
            \raisebox{-.5\height}{\includegraphics[height=\tableImHeight]{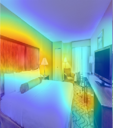}} &
            \raisebox{-.5\height}{\includegraphics[height=\tableImHeight]{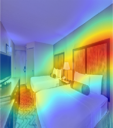}} &
            \raisebox{-.5\height}{\includegraphics[height=\tableImHeight]{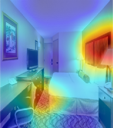}}
        \end{tabular}
        \caption{TraffickCam Hotel Rooms}
    \end{subfigure}
    \caption{Each row shows the regions of the images most representative of the class membership.}
    \label{fig:whatMakesAClass}
\end{figure}

\subsection{Class Similarity}
Using our method, we can discover the most representative components of a class of images. This is a natural extension of class activation maps for classification networks, which visualize the components of an image contribute the most to a particular output label. We generate class activation maps for a given image by summing the pairwise similarity maps with the other images in the same class.
Figure~\ref{fig:whatMakesAClass} shows class similarity visualization for a selection of images from each of our datasets. The visualizations highlight the
portions of the image that most contribute to the similarity of the output feature to those of the images in the same class. For example, in Class 1 of the Google Landmarks dataset, the clock on the building's facade is the most important part in each example image; in Class 2 of the VGG-Faces dataset, the nose and lips are most important; and in Class 3 of the TraffickCam Hotel Rooms dataset, the headboard is most associated with the hotel identity.

\subsection{Object- and Region-Specific Retrieval}
\begin{figure*}
    \centering
    \setlength\tabcolsep{2pt}
    \renewcommand{\arraystretch}{2}
    \begin{subfigure}[b]{\textwidth}
    \begin{tabular}{lc|cc|cc|cc}
       \multicolumn{2}{c|}{Query Image} & \multicolumn{6}{|c}{Top Matches}  \\
        \hline
        Whole Image &
        \raisebox{-.5\height}{\includegraphics[height=\tableImHeight]{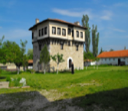}} & 
        \raisebox{-.5\height}{\fcolorbox{green}{green}{\includegraphics[height=\tableImHeight]{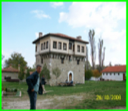}}} & 
        \raisebox{-.5\height}{\includegraphics[height=\tableImHeight]{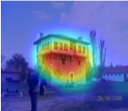}} &
        \raisebox{-.5\height}{\fcolorbox{red}{red}{\includegraphics[height=\tableImHeight]{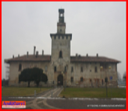}}} & 
        \raisebox{-.5\height}{\includegraphics[height=\tableImHeight]{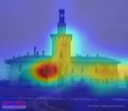}} &
        \raisebox{-.5\height}{\fcolorbox{red}{red}{\includegraphics[height=\tableImHeight]{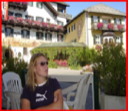}}} & 
        \raisebox{-.5\height}{\includegraphics[height=\tableImHeight]{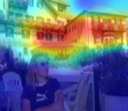}} 
        \\
        Region 1 &
        \raisebox{-.5\height}{\includegraphics[height=\tableImHeight]{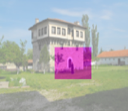}} & 
        \raisebox{-.5\height}{\fcolorbox{green}{green}{\includegraphics[height=\tableImHeight]{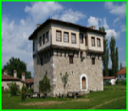}}} & 
        \raisebox{-.5\height}{\includegraphics[height=\tableImHeight]{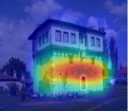}} & 
        \raisebox{-.5\height}{\fcolorbox{red}{red}{\includegraphics[height=\tableImHeight]{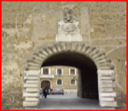}}} & 
        \raisebox{-.5\height}{\includegraphics[height=\tableImHeight]{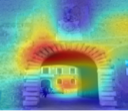}} & 
        \raisebox{-.5\height}{\fcolorbox{red}{red}{\includegraphics[height=\tableImHeight]{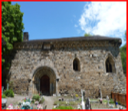}}} & 
        \raisebox{-.5\height}{\includegraphics[height=\tableImHeight]{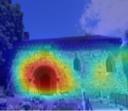}} 
        \\
        Region 2 &
        \raisebox{-.5\height}{\includegraphics[height=\tableImHeight]{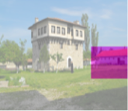}} & 
        \raisebox{-.5\height}{\fcolorbox{red}{red}{\includegraphics[height=\tableImHeight]{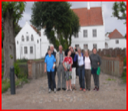}}} & 
        \raisebox{-.5\height}{\includegraphics[height=\tableImHeight]{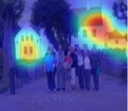}} & 
        \raisebox{-.5\height}{\fcolorbox{red}{red}{\includegraphics[height=\tableImHeight]{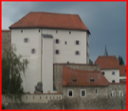}}} & 
        \raisebox{-.5\height}{\includegraphics[height=\tableImHeight]{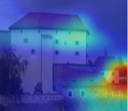}} & 
        \raisebox{-.5\height}{\fcolorbox{red}{red}{\includegraphics[height=\tableImHeight]{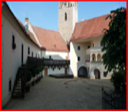}}} & 
        \raisebox{-.5\height}{\includegraphics[height=\tableImHeight]{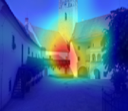}} 
        \end{tabular}
    \caption{Google Landmarks}
    \end{subfigure}
    \begin{subfigure}[b]{\textwidth}
        \begin{tabular}{lc|cc|cc|cc}
        Whole Image &
        \raisebox{-.5\height}{\includegraphics[height=\tableImHeight]{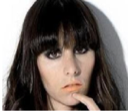}} & 
        \raisebox{-.5\height}{\fcolorbox{green}{green}{\includegraphics[height=\tableImHeight]{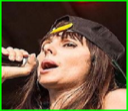}}} & 
        \raisebox{-.5\height}{\includegraphics[height=\tableImHeight]{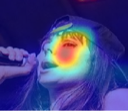}} & 
        \raisebox{-.5\height}{\fcolorbox{red}{red}{\includegraphics[height=\tableImHeight]{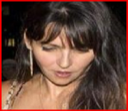}}} & 
        \raisebox{-.5\height}{\includegraphics[height=\tableImHeight]{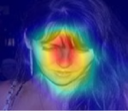}} & 
        \raisebox{-.5\height}{\fcolorbox{red}{red}{\includegraphics[height=\tableImHeight]{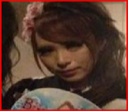}}} & 
        \raisebox{-.5\height}{\includegraphics[height=\tableImHeight]{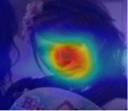}} 
        \\
        Region 1 &
        \raisebox{-.5\height}{\includegraphics[height=\tableImHeight]{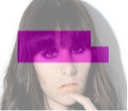}} & 
        \raisebox{-.5\height}{\fcolorbox{red}{red}{\includegraphics[height=\tableImHeight]{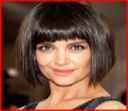}}} & 
        \raisebox{-.5\height}{\includegraphics[height=\tableImHeight]{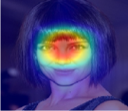}} & 
        \raisebox{-.5\height}{\fcolorbox{green}{green}{\includegraphics[height=\tableImHeight]{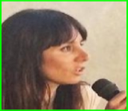}}} & 
        \raisebox{-.5\height}{\includegraphics[height=\tableImHeight]{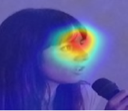}} & 
        \raisebox{-.5\height}{\fcolorbox{red}{red}{\includegraphics[height=\tableImHeight]{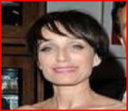}}} & 
        \raisebox{-.5\height}{\includegraphics[height=\tableImHeight]{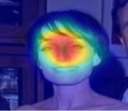}} 
        \\
        Region 2 &
        \raisebox{-.5\height}{\includegraphics[height=\tableImHeight]{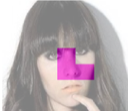}} & 
        \raisebox{-.5\height}{\fcolorbox{red}{red}{\includegraphics[height=\tableImHeight]{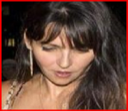}}} & 
        \raisebox{-.5\height}{\includegraphics[height=\tableImHeight]{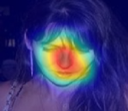}} & 
        \raisebox{-.5\height}{\fcolorbox{red}{red}{\includegraphics[height=\tableImHeight]{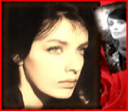}}} & 
        \raisebox{-.5\height}{\includegraphics[height=\tableImHeight]{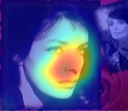}} & 
        \raisebox{-.5\height}{\fcolorbox{red}{red}{\includegraphics[height=\tableImHeight]{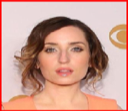}}} & 
        \raisebox{-.5\height}{\includegraphics[height=\tableImHeight]{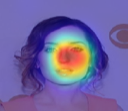}} 
        \end{tabular}
    \caption{VGG-Faces}
    \end{subfigure}
    \begin{subfigure}[b]{\textwidth}
        \begin{tabular}{lc|cc|cc|cc}
        Whole Image &
        \raisebox{-.5\height}{\includegraphics[height=\tableImHeight]{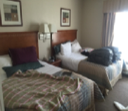}} & 
        \raisebox{-.5\height}{\fcolorbox{green}{green}{\includegraphics[height=\tableImHeight]{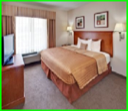}}} & 
        \raisebox{-.5\height}{\includegraphics[height=\tableImHeight]{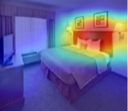}} & 
        \raisebox{-.5\height}{\fcolorbox{red}{red}{\includegraphics[height=\tableImHeight]{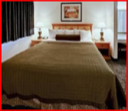}}} & 
        \raisebox{-.5\height}{\includegraphics[height=\tableImHeight]{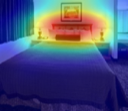}} & 
        \raisebox{-.5\height}{\fcolorbox{red}{red}{\includegraphics[height=\tableImHeight]{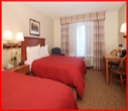}}} & 
        \raisebox{-.5\height}{\includegraphics[height=\tableImHeight]{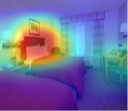}} 
        \\
        Region 1 &
        \raisebox{-.5\height}{\includegraphics[height=\tableImHeight]{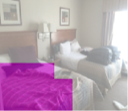}} & 
        \raisebox{-.5\height}{\fcolorbox{red}{red}{\includegraphics[height=\tableImHeight]{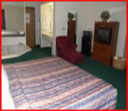}}} & 
        \raisebox{-.5\height}{\includegraphics[height=\tableImHeight]{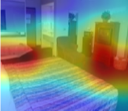}} & 
        \raisebox{-.5\height}{\fcolorbox{red}{red}{\includegraphics[height=\tableImHeight]{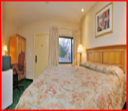}}} & 
        \raisebox{-.5\height}{\includegraphics[height=\tableImHeight]{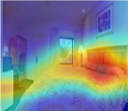}} & 
        \raisebox{-.5\height}{\fcolorbox{red}{red}{\includegraphics[height=\tableImHeight]{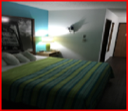}}} & 
        \raisebox{-.5\height}{\includegraphics[height=\tableImHeight]{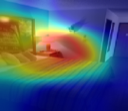}} 
        \\
        Region 2 &
        \raisebox{-.5\height}{\includegraphics[height=\tableImHeight]{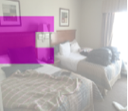}} & 
        \raisebox{-.5\height}{\fcolorbox{red}{red}{\includegraphics[height=\tableImHeight]{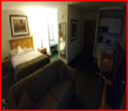}}} & 
        \raisebox{-.5\height}{\includegraphics[height=\tableImHeight]{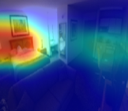}} & 
        \raisebox{-.5\height}{\fcolorbox{red}{red}{\includegraphics[height=\tableImHeight]{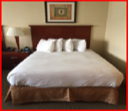}}} & 
        \raisebox{-.5\height}{\includegraphics[height=\tableImHeight]{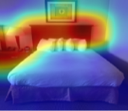}} & 
        \raisebox{-.5\height}{\fcolorbox{red}{red}{\includegraphics[height=\tableImHeight]{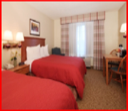}}} & 
        \raisebox{-.5\height}{\includegraphics[height=\tableImHeight]{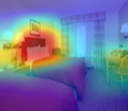}} 
        \end{tabular}
        \caption{TraffickCam}
    \end{subfigure}
    \caption{Object- and Region-Specific Retrieval. We show the most similar image from the three most similar classes when using either the whole image as the query input, or selected sub-regions of the image. This allows for object- or region- specific image retrieval; for example, ``find landmarks with similar archways'', ``find faces with brunette bangs'' or ``find hotels with similar looking bedspreads.'' }
    \label{fig:objSimilarity}
\end{figure*}
The previous experiments highlight the utility of the visualization method
on image retrieval tasks, which consider the entire image as input. The same visualization approach can also support object- or region-specific image retrieval.

We first compute the similarity maps between a query and the database images. Recall that the similarity map for a pair of images sums to the overall similarity score. Therefore, the sum of the values contained within subregions of the similarity map reflect the how much the subregion contributes to the match with the corresponding images. This subselection allows for database images to be sorted based on the contribution to the query image's similarity score at the region of interest. 
Figure~\ref{fig:objSimilarity} shows the most similar image from the three most similar classes for both whole image similarity and also using the highlighted sub-regions of the image. This modification allows for object- or region- specific image retrieval. For example, in the Google Landmarks dataset, searches can be constrained to find landmarks with similar archways or buildings with orange roofs. 

\section{Conclusions}

We present an approach to visualize the image regions responsible for pairwise similarity in an embedding network.  While some previous work visualizes a few of these components, we find that the top few components do not explain most of the similarity.  Our approach explicitly decomposes the entire similarity score between two images and assigns it to the relevant image regions.

We illustrate a number of possible ways to use this visualization tool, exploring differences in networks trained with max pooling and average pooling, illustrating how the focus of a network changes during training, and offering an approach that uses this spatial similarity decomposition to search for matches to objects or sub-regions in an image.

The research area of similarity networks is quite active, exploring variations in the pooling strategies, learned or explicitly pre-defined linear transforms of the pooled feature, and boosting and ensemble strategies.  We have shared code for this project at \url{https://github.com/GWUvision/Similarity-Visualization}, with the goal that these visualizations will provide additional insight about how embeddings are affected by these algorithmic choices.

\bibliographystyle{ieee}
\bibliography{egbib}
\end{document}